\documentclass{article} 
\usepackage{iclr2026_conference,times}

\usepackage{amsmath,amsfonts,bm}

\def\eqref#1{equation~\ref{#1}}

\def\1{\bm{1}}

\DeclareMathAlphabet{\mathsfit}{\encodingdefault}{\sfdefault}{m}{sl}
\SetMathAlphabet{\mathsfit}{bold}{\encodingdefault}{\sfdefault}{bx}{n}

\usepackage[acronym, nonumberlist, nogroupskip]{glossaries}
\usepackage{hyperref}
\usepackage{url}
\usepackage{graphicx}
\usepackage{capt-of}
\usepackage[many]{tcolorbox}
\usepackage{xcolor}
\usepackage{wrapfig}
\usepackage{subcaption}
\usepackage{booktabs}
\newcounter{finding}
\usepackage{enumitem}

\definecolor{findingbox}{RGB}{253, 244, 233}
\definecolor{findingboxboarder}{RGB}{239, 148, 34}
\definecolor{plasmaSize}{RGB}{13, 8, 135}
\definecolor{plasmaPrecision}{RGB}{204, 70, 120}
\definecolor{plasmaGen}{RGB}{252, 165, 10}
\definecolor{plasmaParallel}{RGB}{126, 3, 168}
\definecolor{plasmaKV}{RGB}{225, 100, 98}

\newtcolorbox{findingbox}{
    enhanced,
    breakable,
    colback=findingbox,
    colframe=findingboxboarder,
    boxrule=1.5pt,
    arc=0.25em,
    left=1em,
    right=1em,
    top=1em,
    bottom=0.75em,
    before={\vspace{1em}\refstepcounter{finding}},
    overlay unbroken and first={
        \node[
            fill=findingboxboarder,
            text=white,
            font=\bfseries,
            anchor=west,
            inner xsep=0.75em,
            inner ysep=0.5em,
            rounded corners=0.25em
        ] at ([xshift=0.75em]frame.north west) {Finding~\thefinding};
    }
}

\title{Not All Bits Are Equal: \\
Scale-Dependent Memory Optimization \\
Strategies for Reasoning Models}

\author{
\centerline{
    Junhyuck Kim$^{k}$, Ethan Ewer$^{w}$\thanks{This work was done during an internship at KRAFTON.}, Taehong Moon$^{k}$, Jongho Park$^{b}$, Dimitris Papailiopoulos$^{w,m}$
} \\
\centerline{
    $^{k}$KRAFTON, $^{w}$University of Wisconsin--Madison, $^{b}$UC Berkeley, $^{m}$Microsoft Research
}
}

\newacronym{llm}{LLM}{Large Language Model}
\newacronym{kv}{KV}{Key-Value}

\iclrfinalcopy 
\begin{document}

\maketitle

\begin{abstract}
While 4-bit quantization has emerged as a memory-optimal choice for non-reasoning models and zero-shot tasks across scales, we show that this universal prescription fails for reasoning models, where the KV cache rather than model size can dominate memory. 
Through systematic experiments across 1{,}700 inference scenarios on AIME25 and GPQA-Diamond, we find a \textit{scale-dependent trade-off}: models with an effective size below 8-bit 4B parameters achieve better accuracy by allocating memory to more weights rather than longer generation, while larger models achieve better accuracy by allocating memory to longer generations.
This scale threshold also determines when parallel scaling becomes memory-efficient and whether KV cache eviction outperforms KV quantization. 
Our findings show that memory optimization for LLMs cannot be scale-agnostic, while providing principled guidelines: for small reasoning models, prioritize model capacity over test-time compute, while for larger ones, maximize test-time compute. 
Our results suggest that optimizing reasoning models for deployment requires fundamentally different strategies from those established for non-reasoning models.
\end{abstract}

\begin{figure}[h]
\begin{center}
\includegraphics[width=0.9\linewidth]{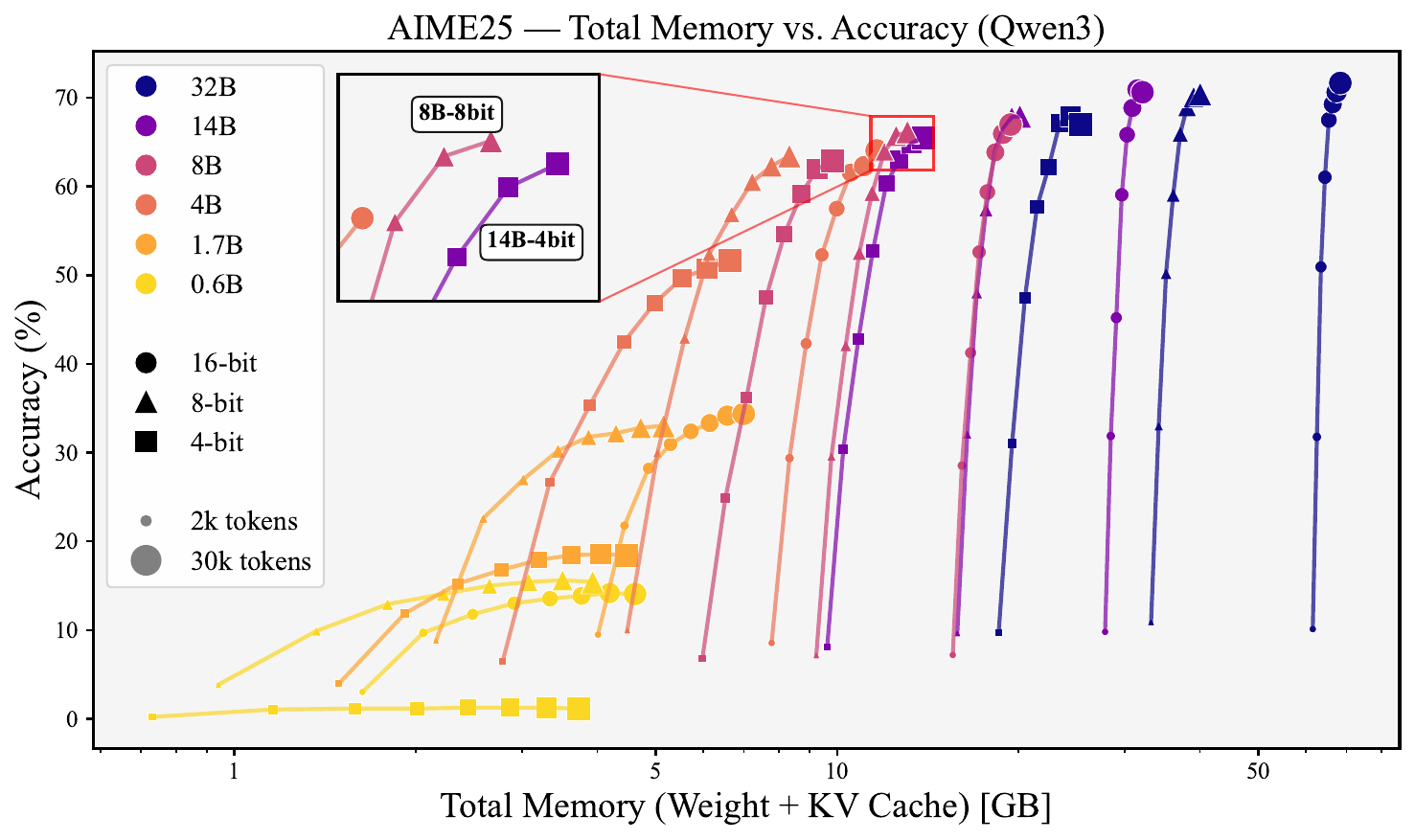}
\end{center}
\caption{
\textbf{Memory vs. Accuracy for serial test-time scaling on AIME25.}
The plot illustrates the trade-off between pass@1 accuracy and total memory (weights + KV cache) for the Qwen3 family.
Model weights are quantized to 4- and 8-bit using GPTQ.
Along each curve, the KV cache grows as the generation length increases via budget forcing.
For models effectively smaller than an 8-bit 4B, increasing the token budget to saturation is memory-inefficient. 
Furthermore, for mathematical reasoning, higher weight precision (8- and 16-bit) proves more memory-efficient than 4-bit.
}
\label{fig:aime25_qwen3_accuracy_total_memory}
\end{figure}

\section{Introduction}
\label{sec:introduction}

Prior memory--performance trade-off studies on \glspl{llm} for non-reasoning models have focused mostly on compressing model weights, since the model weights generally consume far more GPU memory than the \gls{kv} cache~\citep{dettmers2023case,frantar2022gptq,lin2024awq}.
Modern reasoning models, however, generate substantially more tokens, causing the proportionally increasing \gls{kv} cache to become a significant bottleneck.
For instance, a Qwen3-4B model with 4-bit weights occupies 2.49\,GB, but its \gls{kv} cache for a 32k-token generation requires 4.42\,GB (\(\approx 1.8\times\) the weights).
This bottleneck is magnified in batched inference: with model weights amortized, the aggregated \gls{kv} cache becomes the primary memory constraint.
With the \gls{kv} cache becoming a dominant component of memory, it is unclear whether the results established for non-reasoning models still hold for long-generation reasoning tasks.

In this work, we aim to investigate the general principles of memory compression for reasoning models.
In addition to the conventional factors of \textit{\textcolor{plasmaSize}{model size}} and \textit{\textcolor{plasmaPrecision}{weight precision}}, our analysis incorporates three other factors that distinctly affect memory--accuracy trade-offs for reasoning models: \textit{\textcolor{plasmaGen}{generation length}}, \textit{\textcolor{plasmaParallel}{parallel scaling}}, and \textit{\textcolor{plasmaKV}{\gls{kv} cache compression}}.
Overall, we ask the question:
\begin{center}
\textit{Under a fixed memory budget, how should one balance \\
\textnormal{\textbf{model size}}, \textnormal{\textbf{weight precision}}, \textnormal{\textbf{test- and parallel-time compute}}, and \textnormal{\textbf{KV cache compression}} \\
to maximize accuracy on reasoning tasks?}
\end{center}

We conduct an empirical study on the Qwen3 model family (0.6B to 32B)~\citep{yang2025qwen3} across two benchmarks: AIME and GPQA-Diamond. 
Our investigation spans over \textbf{1{,}700} different scenarios, exploring 4-bit and 8-bit GPTQ weight quantization~\citep{frantar2022gptq}, reasoning token budgets from 2k to 30k, parallel scaling via majority voting with up to 16 samples, and two approaches to \gls{kv} cache compression: eviction, using R-KV~\citep{cai2025r} and StreamingLLM~\citep{xiao2023efficient}, and quantization with HQQ~\citep{badri2023hqq}. 
While our findings do not provide specific prescriptions for all tasks or models, we present general principles to consider for memory-efficient reasoning models with minimal loss of accuracy.

\paragraph{Our contributions.}
In Section~\ref{sec:test_time_scaling_with_weight_only_quantization}, we investigate how to allocate memory between model weights and the \gls{kv} cache under serial test-time scaling.
For example, which setting leads to higher accuracy: a 32B 8-bit \gls{llm} with less \gls{kv} cache (\textit{i.e.}, less test-time compute), or a 32B 4-bit \gls{llm} with more \gls{kv} cache?
We find that there is \textit{no} optimal strategy that is universal across scale: for models with effective size (parameters $\times$ bits per weight) below 8-bit 4B (\(\approx 4.2\,\text{GB}\)), allocating more memory to model weights yields larger gains, whereas above this threshold, memory is better spent increasing the test-time budget until performance saturates.

We also discover that the choice of weight precision depends on the nature of the task.
For knowledge-intensive reasoning, 4-bit weight quantization is broadly memory-optimal, consistent with established findings on the effectiveness of 4-bit or lower precision for zero-shot, non-reasoning models~\citep{dettmers2023case,frantar2022gptq,chee2023quip}.
For mathematical reasoning, however, the higher fidelity of 8- or 16-bit model weights with smaller \gls{kv} caches often provides stronger performance, suggesting that intricate computational tasks are more sensitive to loss in precision.

Orthogonal to longer generations, increasing the number of generations can yield substantial gains~\citep{brown2024large}, yet its memory efficiency remains underexplored.
Parallel scaling via majority voting on top of serial scaling introduces another trade-off: a larger \gls{kv} cache proportional to group size for higher accuracy, assuming a batched inference setting.
This strategy is only more memory-efficient than serial scaling for models with an effective size at or above 8-bit 4B. 
Interestingly, for such models, the memory-optimal group size also increases with the total memory budget.

In Section~\ref{sec:test_time_scaling_with_weight_and_kv_cache_compression}, we investigate how KV cache compression affects the memory--accuracy trade-off by considering both KV cache eviction and quantization methods.
Across model sizes and weight precisions, both eviction and quantization advance their Pareto frontiers beyond the baseline without cache compression.
The choice of compression method should be dictated by effective size: eviction offers a better memory trade-off for small models (effective size below 8-bit 4B), while both strategies are competitive for larger models.

Overall, the memory-optimal strategy for reasoning models is \textit{not} universal, but is instead mainly governed by the model's effective size.
We summarize our main empirical findings as follows:
\begin{enumerate}
    \item For models effectively smaller than 8-bit 4B, it is more memory-efficient to allocate memory to more weights than to longer generations, while larger models benefit more from longer generations.
    \item 4-bit weights are broadly memory-optimal for the knowledge-intensive task (GPQA-Diamond), while 8-bit or 16-bit weights are more memory-efficient for the mathematical reasoning task (AIME25).
    \item Parallel scaling only improves the memory--accuracy trade-off for models effectively larger than 8-bit 4B. The memory-optimal group size increases with the memory budget.
    \item Weight quantization alone is not sufficient for memory-optimal reasoning; compressing the \gls{kv} cache leads to more memory-efficient reasoning.
    \item \gls{kv} cache eviction provides a better memory--accuracy trade-off than \gls{kv} cache quantization for models with an effective size smaller than an 8-bit 4B model.
\end{enumerate}

\section{Background}
\label{sec:background}

\paragraph{Weight-only quantization.}
Weight-only post-training quantization replaces full-precision weights with low-bit representations without retraining, reducing memory usage.
Weight-only quantization allows lower bit-widths compared with weight-activation quantization, as it is more robust to quantization error~\citep{yao2023zeroquant}. 
However, weight-only quantization requires dequantization before multiplying with activations, so it does not reduce computational cost during inference. 
Any speedup instead comes from reduced memory movement. 
In this work, we adopt GPTQ~\citep{frantar2022gptq}, a weight-only quantization method that minimizes layer-wise quantization error using a small calibration set and updates weights using inverse-Hessian information.

\paragraph{\gls{kv} cache quantization.}
\gls{kv} cache quantization stores key and value tensors at reduced precision to lower the memory footprint and memory bandwidth during decoding.
Unlike weight-only quantization, \gls{kv} quantization is applied online at inference.
During prefill, the \gls{kv} tensors for the entire input context are quantized and cached in low precision.
During decode, the cached tensors are dequantized on the fly for attention computations.
Prior work conventionally maintains a small full-precision buffer for the most recent tokens, appending new key and value tensors to this buffer during decoding.
In this work, we use per-channel symmetric quantization of both keys and values with an HQQ backend~\citep{badri2023hqq}, a fast, calibration-free quantization method that is particularly well-suited for online \gls{kv} cache quantization.

\paragraph{\gls{kv} cache eviction.}
On the other hand, \gls{kv} cache eviction has also emerged as a critical optimization strategy, reducing \gls{kv} cache size and the cost of attention computation.
Specifically for reasoning models, we consider dynamic eviction policies that continuously evict the \gls{kv} cache during decoding.
Early work, such as StreamingLLM~\citep{xiao2023efficient}, employs a sliding-window mechanism that preserves the most recent key and value tensors, in addition to the initial sequence tokens known as the attention sink.
More recently, R-KV~\citep{cai2025r} proposes redundancy-aware selection for reasoning models: it estimates token importance and redundancy during decoding and jointly selects non-redundant, informative tokens to retain, reporting near-baseline accuracy with a small fraction of the \gls{kv} cache.
In Section~\ref{sec:test_time_scaling_with_weight_and_kv_cache_compression}, we study how these eviction policies, together with \gls{kv} cache quantization, shift the trade-off frontiers.

\paragraph{Test-time scaling.}
We scope this work to test-time scaling methods that do not rely on external models such as verifiers or process reward models.
Reasoning models are typically trained to produce an extended chain-of-thought, continuing generation with planning and reflection to improve performance~\citep{guo2025deepseek,jaech2024openai,yang2025qwen3}. 
We refer to this as serial scaling.
\citet{muennighoff2025s1} introduces budget forcing to scale serial responses beyond the model's natural length for higher accuracy. 
When the model attempts to stop, a short cue is appended to continue decoding to a specified token budget.
Another line of work, parallel scaling, generates multiple independent reasoning trajectories~\citep{brown2024large}.
In its simplest form, without any external model, majority voting selects the final answer as the most frequent among the independently sampled outputs~\citep{wang2022self}.
Further related work is discussed in Section~\ref{sec:related_work}.

\section{Experimental Setup}
\label{sec:experimental_setup}

We systematically explore the memory--accuracy trade-offs by measuring how accuracy and memory footprints are affected by five key factors: the number of parameters ($N$), weight precision ($P_W$), test-time token budget ($T$), sampling group size ($G$, with $G>1$ indicating multiple samples for majority voting), and \gls{kv} cache compression strategy ($\pi_{\mathrm{kv}}$, e.g., eviction or quantization).

The memory cost is given by
\[
M \;=\; M_{\mathrm{weights}}(N, P_W) \; +\; M_{\mathrm{kv}}(N, \pi_{\mathrm{kv}}, T, G),
\]
where $M_{\mathrm{weights}}$ is the memory footprint of the weights, roughly proportional to $N \cdot P_W$.
Note that throughout the paper, we use model \textit{size} to refer to the number of parameters $N$ and \textit{effective size} or \textit{scale} to refer to the memory footprint of the weights, $M_{\mathrm{weights}}$.
$M_{\mathrm{kv}}$ is the \gls{kv} cache memory, which is roughly proportional to $N$, $G$, and $T$, except when $\pi_{\mathrm{kv}} = \text{eviction}$, where the cost becomes constant beyond a certain token budget.
Please refer to Appendix~\ref{sec:memory_equations} for the exact memory cost equations and Table~\ref{tab:model_specs} for model-specific values.

\begin{table}[h]
\centering
\caption{\textbf{Memory footprints of evaluated models.}}
\label{tab:model_specs}
\resizebox{\textwidth}{!}{%
\begin{tabular}{l|ccc|cccc}
\toprule
\textbf{Model} & \multicolumn{3}{c|}{\textbf{Model Weight (GB)}} & \multicolumn{4}{c}{\textbf{KV Cache (GB)}} \\
 & \textbf{4-bit} & \textbf{8-bit} & \textbf{16-bit} & \textbf{2k tokens} & \textbf{18k tokens} & \textbf{30k tokens} & \textbf{30k tokens $\times$ 16 samples} \\
\midrule
Qwen3-0.6B & 0.50 & 0.71 & 1.40 & 0.21 & 1.92 & 3.20 & 51.27 \\
Qwen3-1.7B & 1.26 & 1.93 & 3.78 & 0.21 & 1.92 & 3.20 & 51.27 \\
Qwen3-4B & 2.49 & 4.19 & 7.49 & 0.27 & 2.47 & 4.12 & 65.91 \\
Qwen3-8B & 5.68 & 8.94 & 15.26 & 0.27 & 2.47 & 4.12 & 65.91 \\
Qwen3-14B & 9.30 & 15.50 & 27.51 & 0.31 & 2.75 & 4.58 & 73.24 \\
Qwen3-32B & 18.01 & 32.66 & 61.02 & 0.49 & 4.39 & 7.32 & 117.19 \\
\bottomrule
\end{tabular}%
}
\end{table}

\paragraph{Models.}
We experiment with the Qwen3 model family~\citep{yang2025qwen3}, which ranges from 0.6B to 32B parameters, offering a wide range of model sizes and thus making it well-suited for a fine-grained systematic study across scales.

\paragraph{Tasks.}
Experiments are conducted on challenging benchmarks representing complementary difficulty profiles.
AIME25~\citep{aime2025problems} is a competition-level mathematical benchmark that stresses multi-step reasoning.
In contrast, GPQA-Diamond~\citep{rein2024gpqa} emphasizes scientific knowledge and integrated reasoning across domains such as chemistry, biology, and physics~\citep{li2025system}.

\paragraph{Inference details.}
Unless otherwise specified, we report accuracy averaged over 32 generations per instance, sampling with temperature $0.6$.
Following \citet{muennighoff2025s1}, for serial scaling with budget forcing, if generation terminates earlier than the desired token budget, we replace the end-of-sequence token with the prompt \texttt{ Wait} and continue decoding until the target budget is reached.
When the desired budget is met, we inject the prompt \texttt{**Final Answer**\textbackslash n\textbackslash\textbackslash boxed\{}.
We evaluate token budgets from 2k to 30k in 4k increments.
Our code is available at \url{https://github.com/krafton-ai/not-all-bits-are-equal}.

\section{Test-Time Scaling with Weight-Only Quantization}
\label{sec:test_time_scaling_with_weight_only_quantization}

\begin{center}
    \textit{
    When aiming for the best performance under limited memory, how should memory be allocated between model weights and KV cache? 
    Additionally, when allocating space for model weights, is it better to use more parameters at lower precision or fewer parameters at higher precision?
    }
\end{center}

To answer these questions, we study test-time scaling across different model sizes ($N$) and weight precisions ($P_W \in \{4,8,16\}$) by varying the test-time token budget ($T$). 
We use GPTQ to quantize models to 4- and 8-bit precision.
For this analysis, we fix $\pi_{\mathrm{kv}}$ to keep all cache entries (no eviction, at full precision) and first present results for a sampling group size of $G=1$. 
We later discuss parallel scaling with $G>1$ and other $\pi_{\mathrm{kv}}$ policies.

\begin{figure}[h]
\begin{center}
\includegraphics[width=\linewidth]{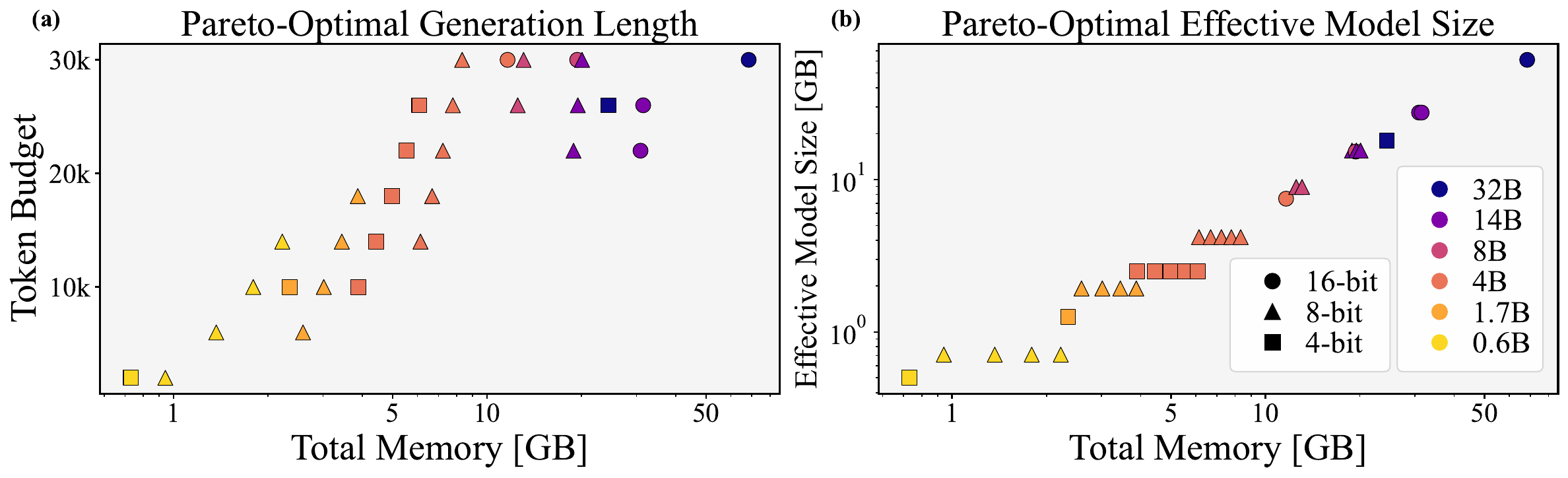}
\end{center}
\caption{
\textbf{Composition of Pareto-optimal configurations (AIME25, Qwen3).} 
The token budget (a) and effective model size (b) are plotted against the total memory budget for configurations on the Pareto frontier from Figure~\ref{fig:aime25_qwen3_accuracy_total_memory}. 
The plots illustrate a strategic shift: at lower memory budgets (\textless{}10\,GB), increasing effective model size is memory-efficient, whereas at higher budgets, increasing the token budget becomes the dominant strategy for improving performance.}
\label{fig:aime25_qwen3_optimal_configs}
\end{figure}

Figure~\ref{fig:aime25_qwen3_accuracy_total_memory} reveals the Pareto frontier for accuracy versus total memory under serial scaling with a full-precision \gls{kv} cache. 
Analyzing the configurations that lie on this frontier provides practical recommendations for optimizing model selection, weight precision, and test-time budgets within fixed memory constraints:

\paragraph{For models effectively smaller than 8-bit 4B, memory is better spent on increasing the effective model size rather than increasing the test-time budget until saturation.}
While extending the generation budget of a small model is often viewed as a way to trade higher latency for lower memory usage compared to using a large model, our analysis reveals that this is a false economy.
In fact, for models effectively smaller than 8-bit 4B, this strategy is often suboptimal in total memory.
Figure~\ref{fig:aime25_qwen3_optimal_configs} shows that for memory budgets below 8\,GB, the Pareto frontier is advanced primarily by increasing model size, not the token budget. 
For instance, the 1.7B model in 8-bit with a 6k token budget outperforms the 0.6B model in 8-bit with an 18k token budget.
Similarly, the 4B model in 4-bit with a 10k token budget surpasses the 1.7B model in 8-bit with an 18k token budget, demonstrating that choosing a model with a larger effective size is better under a similar memory budget. 
As our latency analysis confirms (Section~\ref{sec:latency_throughput_analysis}), these configurations with larger effective sizes are also faster because end-to-end latency is dominated by the token budget, making the choice to increase the model's effective size strictly dominant.

\paragraph{For large models with an effective size at or above 8-bit 4B, memory is more efficiently used when increasing the test-time budget until performance saturates.}
In direct contrast to the strategy for small models, extending the generation budget is a more memory-efficient way to improve accuracy for large models.
This strategic shift is clearly illustrated in Figure~\ref{fig:aime25_qwen3_optimal_configs}, where for memory budgets larger than 10\,GB, the best-performing configurations on the Pareto frontier consistently feature token budgets above 20k. 
In this regime, increasing the token budget becomes the dominant method for improving accuracy.

\begin{findingbox}
    The memory-efficient allocation strategy between model weights and KV cache is scale-dependent.
    For models effectively smaller than 8-bit 4B, memory is more efficiently allocated to increasing the effective model size. 
    For models at or above this threshold, it becomes more memory-efficient to increase the test-time budget until performance saturates.
 \end{findingbox}

\begin{figure}[h]
\begin{center}
\includegraphics[width=\linewidth]{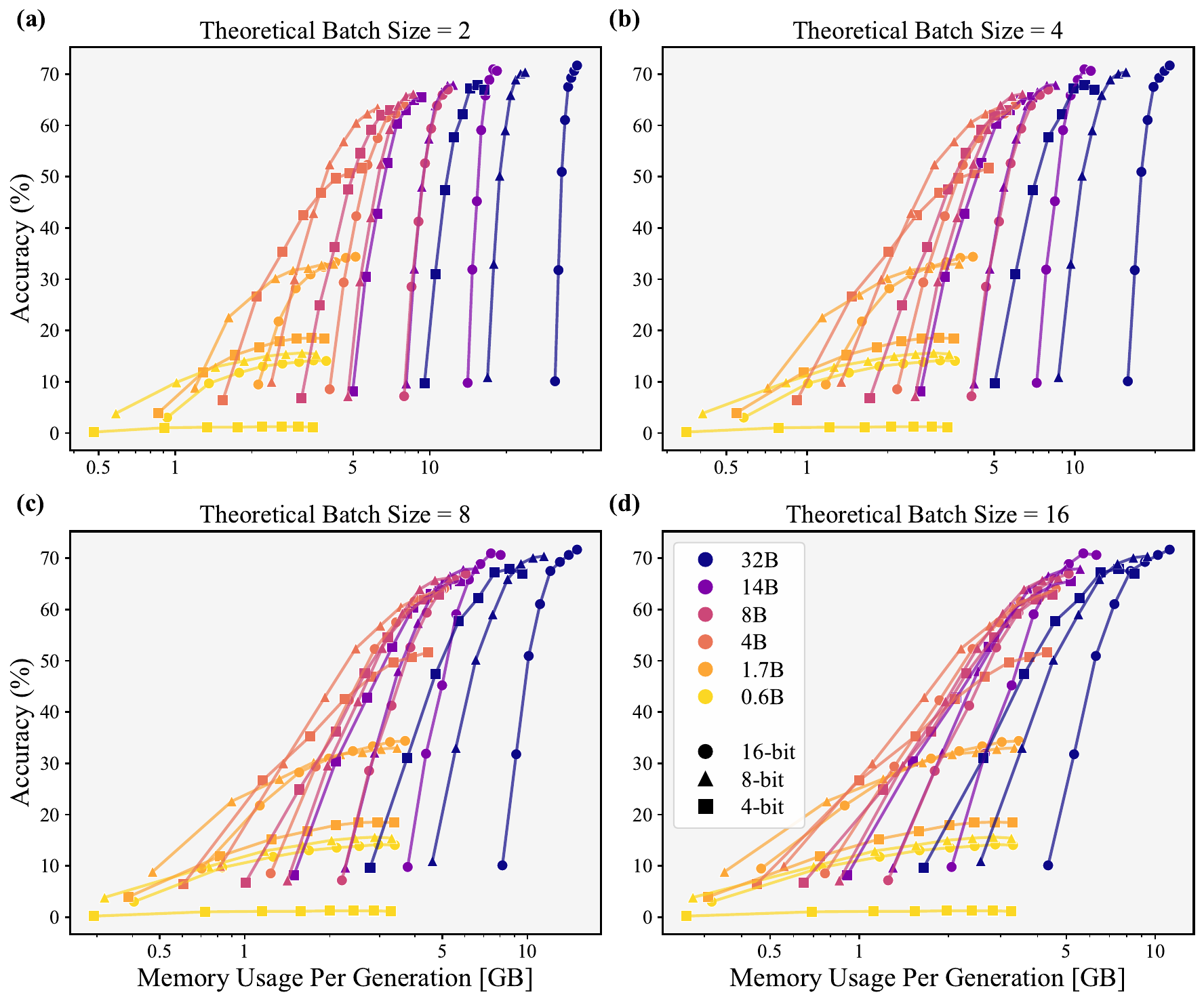}
\end{center}
\caption{
\textbf{Memory vs. Accuracy under different theoretical batch sizes (AIME25, Qwen3).}
Each subplot shows memory-per-generation vs. accuracy for different theoretical batch sizes, where model weight memory is amortized across concurrent generations. 
The Pareto frontier shifts as batch size increases, revealing how model weight amortization affects the optimal memory allocation strategy.
}
\label{fig:aime25_qwen3_theoretical_batch_size}
\end{figure}

While our analysis mainly assumes a scenario where each inference instance uses the entire model and \gls{kv} cache, in practice, model weights can be amortized across multiple concurrent generations, fundamentally changing the memory dynamics.
Figure~\ref{fig:aime25_qwen3_theoretical_batch_size} examines how memory--accuracy trade-offs change when model weights are shared across multiple concurrent generations.
As the theoretical batch size increases, the benefit of smaller model weights diminishes because weight costs are amortized across more generations.
We find that the 0.6B model never appears on the Pareto frontier at a theoretical batch size of 16.
The 8B and 14B models with 4-bit and 8-bit weight precision and the 4B model with 8-bit and 16-bit precision demonstrate favorable trade-offs in the 1--4\,GB memory-per-generation region when the theoretical batch size is 16.
Notably, the 8-bit 4B model consistently lies on the Pareto frontier for the 1--2\,GB region.

\begin{figure}[h]
\begin{center}
\begin{minipage}[t]{0.48\linewidth}
\includegraphics[width=\linewidth]{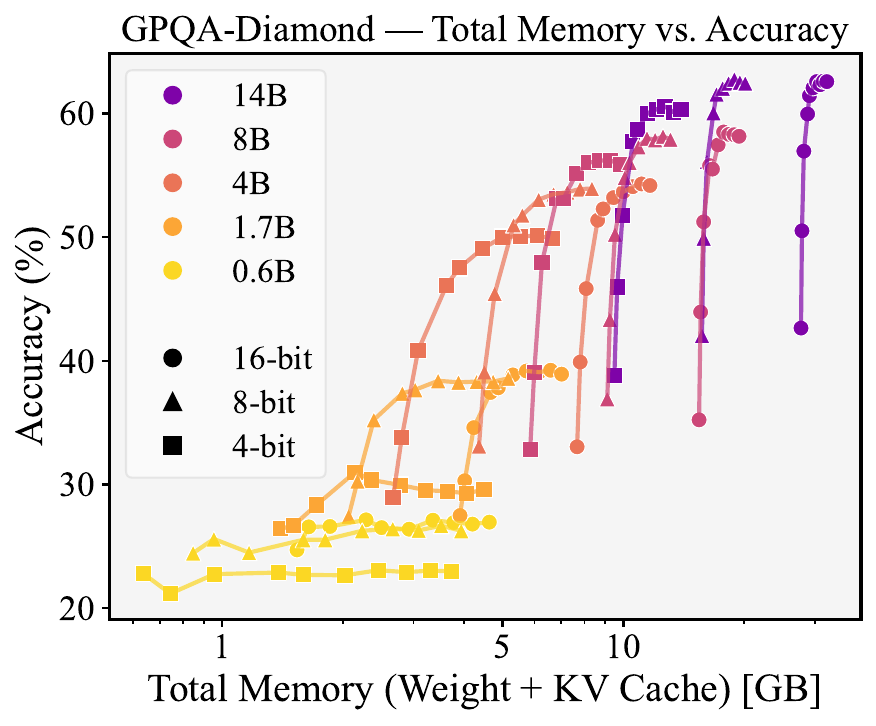}
\captionof{figure}{
\textbf{Memory vs. Accuracy on GPQA-Diamond (Qwen3).} 
The memory--accuracy trade-off for serial scaling on GPQA-Diamond.
Total memory is the sum of model weights and KV cache. 
Points along each curve represent increasing token budgets.
4-bit weights are broadly memory-optimal for knowledge-intensive tasks.
}
\label{fig:gpqad_qwen3_accuracy_total_memory}
\end{minipage}\hfill
\begin{minipage}[t]{0.48\linewidth}
\includegraphics[width=\linewidth]{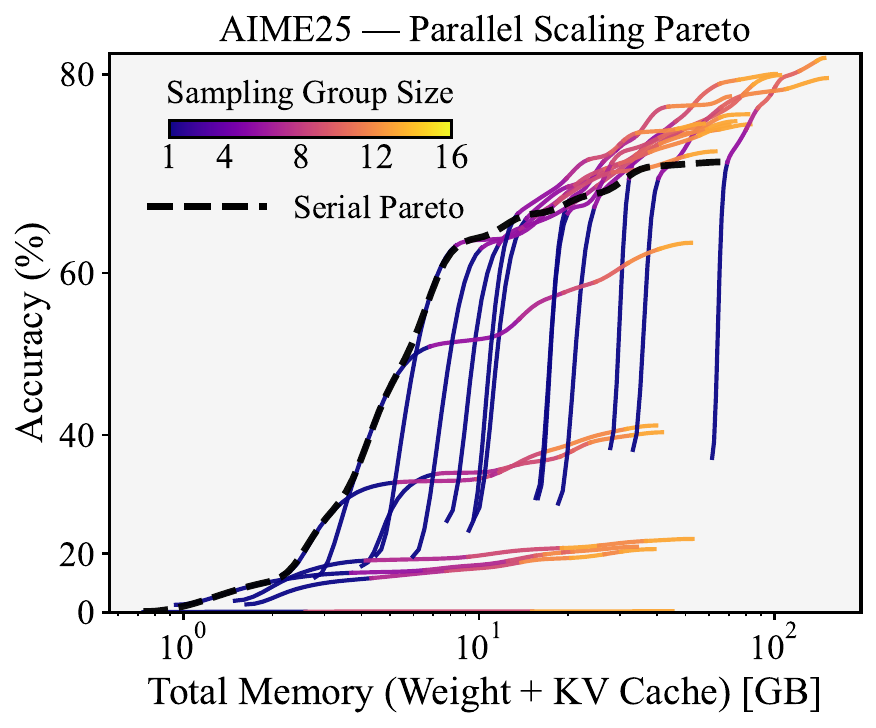}
\captionof{figure}{
\textbf{Effect of parallel scaling on the Pareto frontier (Qwen3).} 
Each colored curve represents the Pareto frontier for a specific model size and weight precision, obtained by increasing the sampling group size, $G$. 
The Pareto frontier for serial scaling ($G=1$) across all models is shown as a dotted line.
Parallel scaling is only effective for large models.
}
\label{fig:aime25_qwen3_parallel_scaling_pareto}
\end{minipage}
\end{center}
\end{figure}

\paragraph{The memory-optimal weight precision is task- and size-dependent.}
Returning to the trade-offs in a single-batch inference setting (Figures~\ref{fig:aime25_qwen3_accuracy_total_memory} and \ref{fig:aime25_qwen3_optimal_configs}), our findings show that for mathematical reasoning tasks, 4-bit weight quantization is consistently memory-inefficient.
On the AIME25 benchmark, 8-bit is memory-optimal for small models ($N \in \{0.6\text{B}, 1.7\text{B}\}$), as the performance gains from reallocating memory saved by 4-bit quantization to a larger token budget are insufficient to compensate for the accuracy loss.
This inefficiency of 4-bit persists at larger $N$, where 8-bit and 16-bit configurations achieve higher accuracy at comparable memory. 
This is shown in Figure~\ref{fig:aime25_qwen3_optimal_configs} (b), where 8- or 16-bit weights are most often memory-optimal along the frontier for memory budgets larger than 6\,GB.
Notably, the 8B model in 8-bit consistently outperforms the 14B model in 4-bit (Figure~\ref{fig:aime25_qwen3_accuracy_total_memory}), and the 32B model in 4-bit is strictly dominated by both the 14B model in 8-bit and the 8B model in 16-bit.
Such findings are in direct contrast to~\citet{dettmers2023case}.
However, we do find that for knowledge-intensive tasks, 4-bit quantization is broadly memory-optimal.
As shown in Figure~\ref{fig:gpqad_qwen3_accuracy_total_memory} for GPQA-Diamond, the frontier shifts to favor lower precision.
This suggests that different task types place different demands on model parameters. 
Mathematical reasoning may rely on numerical precision within the weights, which is damaged by aggressive 4-bit quantization. 
On the other hand, knowledge-intensive tasks prioritize maximizing the number of parameters to increase knowledge capacity, making large 4-bit models more memory-efficient.

\begin{findingbox}
   For knowledge-intensive tasks, 4-bit is broadly memory-optimal. 
   For mathematical reasoning tasks, higher precision is required. 
   8-bit is memory-optimal for small models ($N \in \{0.6\text{B}, 1.7\text{B}\}$), while both 8-bit and 16-bit are competitive at larger numbers of parameters.
\end{findingbox}

In addition to serial scaling by increasing the token budget, we can introduce a parallel scaling axis by increasing the sampling group size ($G$).
Assuming a batched inference setting, the \gls{kv} cache grows with $G$, in exchange for higher accuracy.
This raises another key question:

\begin{center}
    \textit{
    When is it more memory-efficient to allocate memory to parallel samples, versus to a larger effective model size or a longer generation length?
    }
\end{center}

\paragraph{The effectiveness of parallel scaling is scale-dependent.}
For systematic evaluation, we use budget forcing to control the token budget for each of the $G$ parallel samples and use majority voting to select the final answer. 
Figure~\ref{fig:aime25_qwen3_parallel_scaling_pareto} shows how parallel scaling affects the memory--accuracy trade-off. 
The dotted line marks the Pareto frontier from serial scaling alone. 
Each colored curve represents the frontier for a specific model configuration as the group size, $G$, is increased (see Appendix~\ref{sec:detailed_results_for_parallel_scaling}, Figure~\ref{fig:aime25_qwen3_parallel_scaling_individual} for a per-model breakdown).
For models effectively smaller than 8-bit 4B, parallel scaling is memory-inefficient, as its configurations lie below the frontier established by serial scaling alone.
However, for large models, parallel scaling improves the trade-off, and the memory-optimal group size $G$ on the global Pareto frontier increases with the memory budget. 
While group sizes of $4 \le G < 8$ are memory-optimal in the 16.4--28.9\,GB range; for budgets above 28.9\,GB, the frontier is pushed by even larger groups ($G \ge 8$).

\begin{findingbox}
   For models effectively smaller than 8-bit 4B, serial scaling alone provides a better memory--accuracy trade-off than parallel scaling. 
   For models effectively larger than this threshold, parallel scaling improves the trade-off, and the memory-optimal group size $G$ on the global Pareto frontier increases with the memory budget.
\end{findingbox}

\subsection{Latency and Throughput Analysis}
\label{sec:latency_throughput_analysis}

While we focus primarily on memory--accuracy trade-offs, latency and throughput can be important practical considerations as well.
In this section, we analyze how model size, weight precision, and generation length affect both metrics.

\paragraph{Experimental setup.}
All measurements are performed on a single NVIDIA A100 80\,GB GPU using the vLLM framework~\citep{kwon2023efficient} with FlashAttention~\citep{dao2023flashattention} as the attention backend.
To measure throughput for a given token budget, we sweep a range of batch sizes and record the highest batch size that completes successfully without out-of-memory errors or \gls{kv} cache preemption.

\begin{figure}[h]
\begin{center}
\begin{minipage}[t]{0.48\linewidth}
\includegraphics[width=\linewidth]{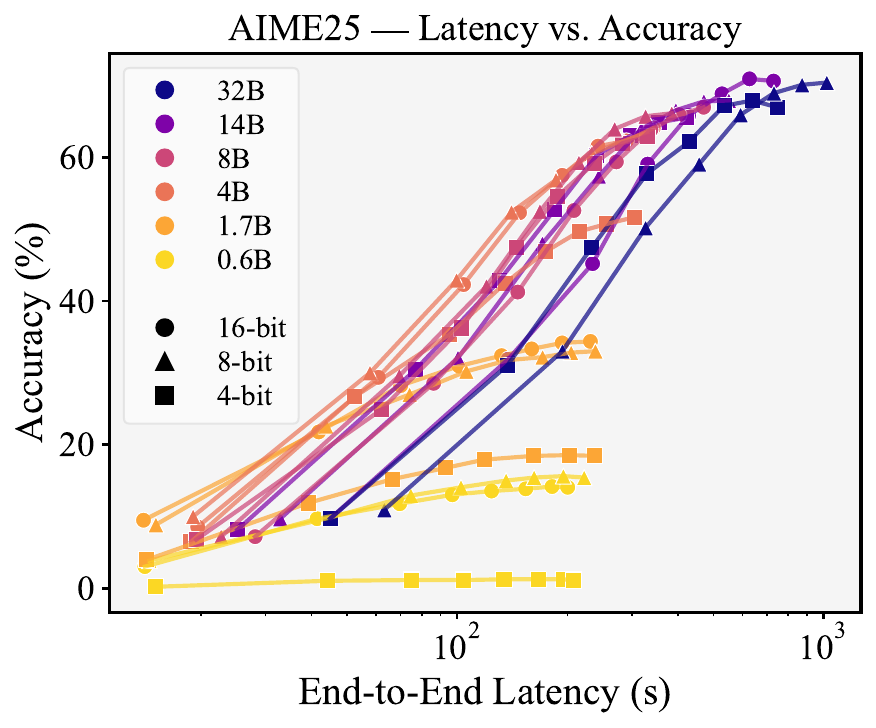}
\caption{
\textbf{Latency vs. Accuracy trade-offs (AIME25, Qwen3).}
Each curve shows end-to-end latency vs. accuracy for different model sizes and weight precisions with increasing generation length.
Generation length emerges as the dominant factor in determining latency, with weight quantization providing more noticeable speedups for large models (14B, 32B).
}
\label{fig:latency}
\end{minipage}\hfill
\begin{minipage}[t]{0.48\linewidth}
\includegraphics[width=\linewidth]{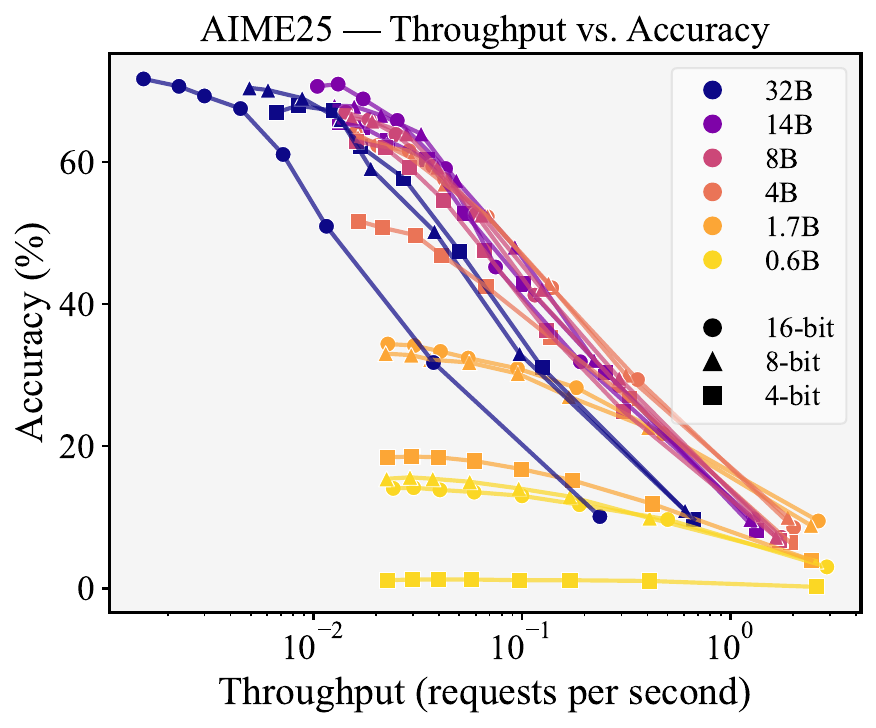}
\caption{
\textbf{Throughput vs. Accuracy trade-offs (AIME25, Qwen3).}
Each point represents the maximum throughput (requests per second) vs. accuracy under 80\,GB VRAM constraints with increasing generation length.
While small models can achieve higher batch sizes, the frontier is dominated by configurations that balance model capability with generation efficiency.
}
\label{fig:throughput}
\end{minipage}
\end{center}
\end{figure}

We show in Figure~\ref{fig:latency} that generation length is the dominant factor determining end-to-end latency across all model configurations.
The benefit of weight quantization on latency due to reduced memory movement costs is modest for small models (up to 8B) but becomes noticeable for larger models (14B, 32B).
For instance, the 14B model takes 137.7 seconds to generate 6k tokens at 16-bit precision, while the 4-bit variant generates 10k tokens in 130.1 seconds.
The overall trend for throughput, shown in Figure~\ref{fig:throughput}, is similar.
   
For both latency and throughput, the 4B model with 8-bit and 16-bit precisions consistently demonstrates the strongest speed--accuracy trade-off. 
Crucially, 4-bit precision is never on the Pareto frontier for any model size, suggesting that for speed-critical applications like reinforcement learning rollouts, higher weight precisions, such as 8-bit, may be the optimal choice~\citep{yao2025flashrl}. 
The trade-offs become less favorable at the extremes of the scale: small size models (0.6B, 1.7B) achieve extreme batch sizes up to 160 and 170, respectively, yielding throughput of 2.9 and 2.64 requests per second with 2k-token generations, but their accuracy is fundamentally limited. 
Conversely, the 32B model performs poorly on throughput due to its slow generation speed and large memory footprint, which restricts batching under an 80\,GB VRAM constraint.

\begin{findingbox}
   For both latency and throughput, 4-bit precision is never on the Pareto frontier, as higher precisions (8-bit and 16-bit) consistently provide a better trade-off between accuracy and speed.
\end{findingbox}

\section{Test-Time Scaling with Weight and KV Cache Compression}
\label{sec:test_time_scaling_with_weight_and_kv_cache_compression}

Our analysis so far shows that while allocating more tokens generally improves accuracy, it is not always memory-efficient, especially for effectively small models where the \gls{kv} cache can dominate total memory.
While compressing the \gls{kv} cache via quantization or eviction can reduce this footprint, it comes at a potential accuracy cost. 
This raises the following question:

\begin{center}
\textit{
    How do KV cache compression strategies---eviction and quantization---alter the overall memory--accuracy trade-off, and which approach leads to stronger reasoning?
    }
\end{center}

To answer this, we evaluate both compression strategies across model sizes and weight precisions. 
For eviction, we use R-KV with target \gls{kv} budgets of 2k, 4k, and 8k tokens.
For \gls{kv} cache quantization, we use symmetric per-channel quantization to 2-, 4-, and 8-bit precisions with a group size of 64 and a full-precision residual buffer of 128 tokens.
The results are averaged over 8 generations per instance.
We first show that both methods are broadly beneficial and then provide a detailed analysis to determine which strategy is optimal under different conditions.

\begin{figure}[h]
\begin{center}
\includegraphics[width=0.65\linewidth]{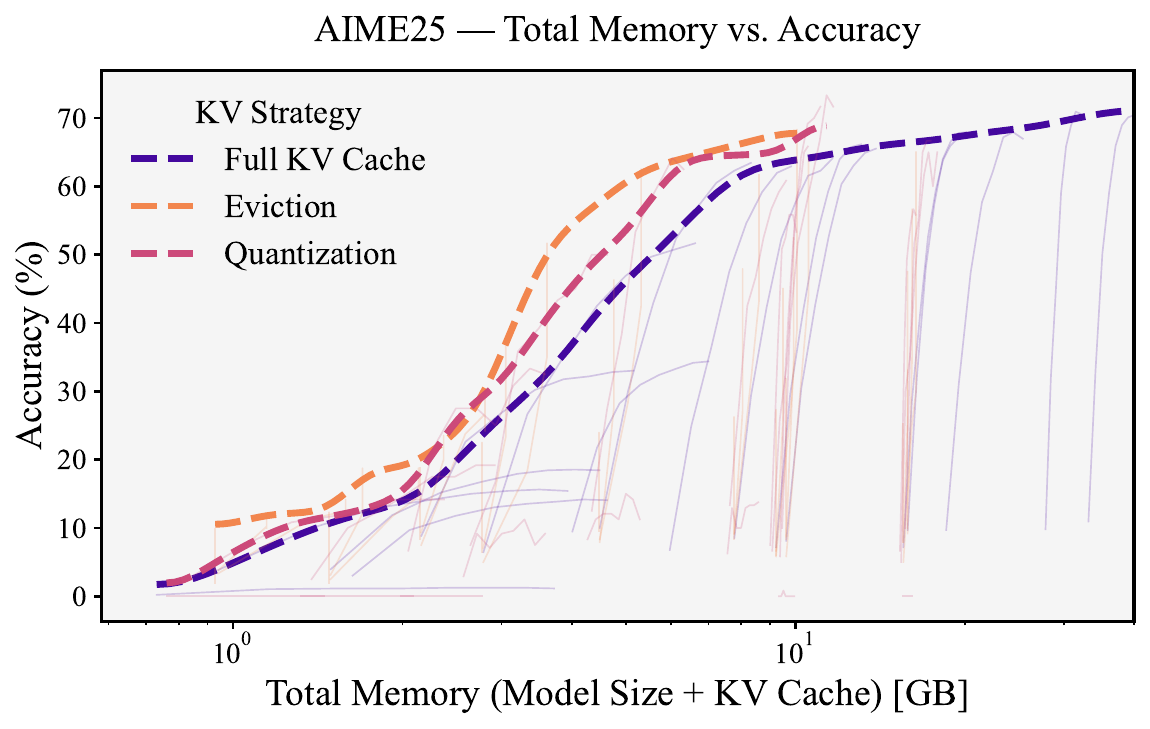}
\end{center}
\caption{
\textbf{Memory vs. Accuracy by \gls{kv} cache compression strategy (AIME25, Qwen3).}
The plot shows the Pareto frontiers of \gls{kv} cache compression across model sizes and weight precisions under serial scaling with budget forcing.
Eviction uses R-KV with token budgets of 2k, 4k, and 8k.
Quantization is symmetric per-channel (group size 64) at 2-, 4-, and 8-bit. 
Faint background lines show curves for individual (model size, weight precision, \gls{kv} strategy) configurations.
Both compression strategies consistently improve the memory--accuracy trade-off.
}
\label{fig:aime25_qwen3_accuracy_total_memory_kv_strategy}
\end{figure}

\paragraph{\gls{kv} cache eviction and quantization consistently advance the Pareto frontier across all tested model sizes and weight precisions.}
Our first key finding, illustrated in Figure~\ref{fig:aime25_qwen3_accuracy_total_memory_kv_strategy}, is that the aggregate Pareto frontiers for both quantization and eviction decisively advance beyond a baseline without compression for models with 4-bit, 8-bit, and 16-bit weights.
This improvement demonstrates that these strategies enable either higher accuracy at the same memory budget or the same accuracy at a lower memory cost, regardless of the model weight precision. 
The benefits are especially pronounced in the low-memory regime below 10\,GB, where smaller models are most constrained by the \gls{kv} cache.
This indicates that even when model weights are aggressively compressed, the \gls{kv} cache contains significant redundancies that can be exploited. 
Our results, therefore, establish \gls{kv} cache compression as an essential and broadly beneficial strategy for the memory-efficient deployment of reasoning models.

\begin{findingbox}
   Weight quantization alone is not sufficient for memory-optimal reasoning.
   \gls{kv} cache compression advances the memory--accuracy frontier across all weight precisions. 
\end{findingbox}

\begin{figure}[h]
\begin{center}
\includegraphics[width=1.0\linewidth]{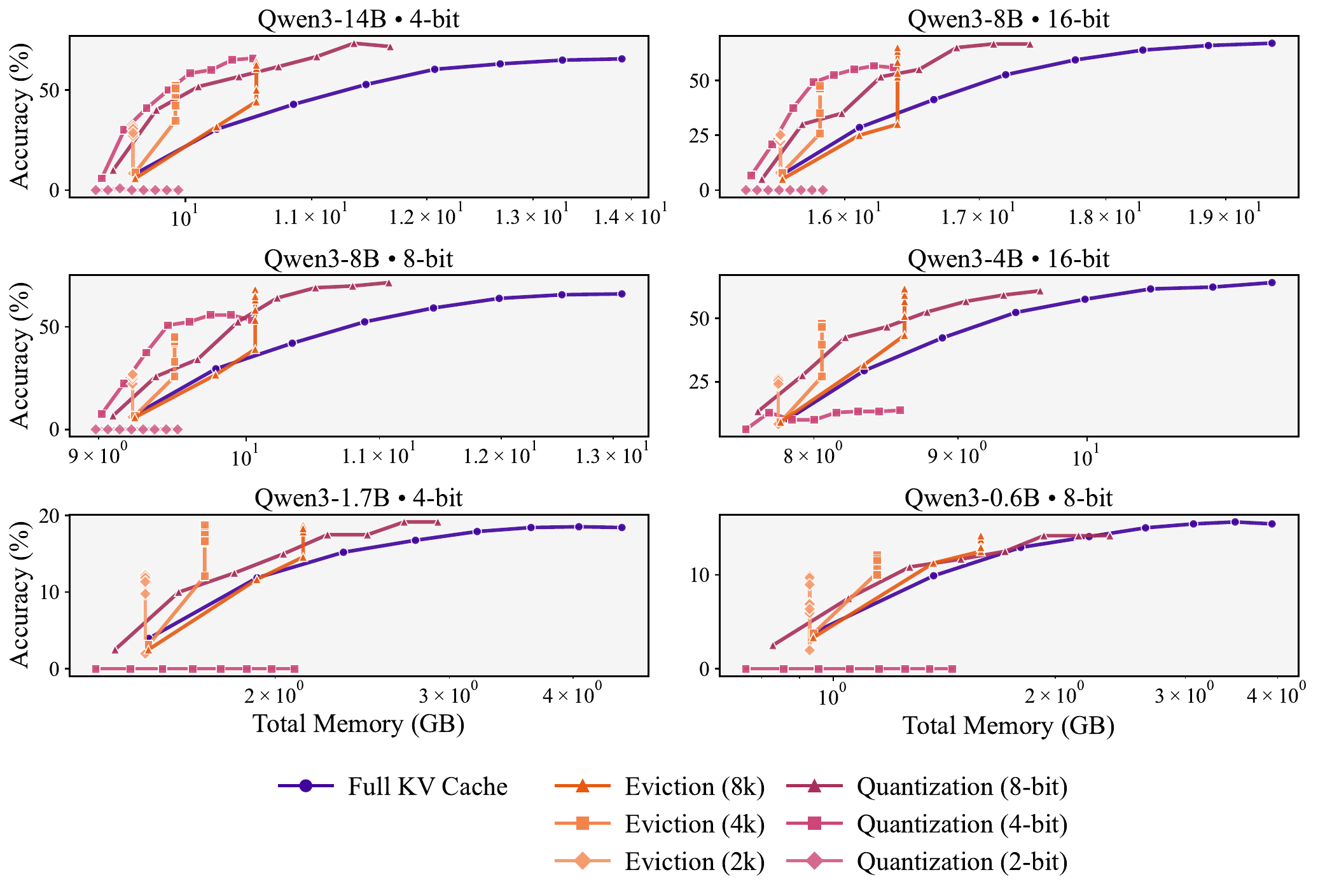}
\end{center}
\caption{
\textbf{Per-model Memory vs. Accuracy by \gls{kv} cache strategy (AIME25).}
Each plot illustrates the memory--accuracy trade-off for a single model size and weight precision, comparing a full \gls{kv} cache baseline against R-KV eviction and symmetric per-channel quantization. 
Points along each curve represent an increasing number of processed tokens via budget forcing.
}
\label{fig:aime25_qwen3_accuracy_total_memory_kv_strategy_individual}
\end{figure}

Having established that \gls{kv} cache compression is broadly beneficial, we now analyze which compression strategy, quantization or eviction, is preferable for a given model size $N$ and weight precision $P_W$. 
Figure~\ref{fig:aime25_qwen3_accuracy_total_memory_kv_strategy_individual} shows the resulting memory--accuracy trade-offs, where each strategy shapes the curves differently. 
Quantization reduces the memory cost per token, shifting the curves leftward, typically with some accuracy degradation. 
Eviction, in contrast, enforces a fixed memory ceiling for the \gls{kv} cache, resulting in characteristic vertical curves where accuracy improves while memory usage remains constant.

\paragraph{Eviction is more effective than quantization for small models.}
For models with an effective size smaller than an 8-bit 8B model, eviction consistently provides the best memory--accuracy trade-off.
As shown in Figure~\ref{fig:aime25_qwen3_accuracy_total_memory_kv_strategy_individual} for the full-precision 4B model, eviction with an 8k token budget maintains near-lossless in maximum accuracy while substantially reducing total memory.
This observation holds across all weight precisions for the 4B model (see Appendix~\ref{sec:detailed_results_for_kv_cache_compression}, Figure~\ref{fig:aime25_qwen3_kv_strategy_all_configs_part2} for these results).
In contrast, aggressive 4-bit \gls{kv} cache quantization causes a significant drop in accuracy at these small effective sizes.
This suggests that effectively small models are more sensitive to the numerical errors introduced by quantization, whereas eviction preserves the full precision of a smaller, more critical set of tokens. 
For instance, on the 1.7B model with 4-bit weight precision, eviction achieves the best memory trade-off while maintaining high accuracy, whereas an 8-bit quantized \gls{kv} cache, while effective, requires significantly more memory to reach a similar performance level.

\paragraph{Quantization becomes competitive with eviction for large models.}
For models with an effective size larger than an 8-bit 8B model, the clear advantage of eviction diminishes as quantization becomes a highly competitive strategy. 
On the 8B model with 16-bit weights, for example, quantization and eviction achieve comparable memory--accuracy trade-offs. 
While 4-bit \gls{kv} cache quantization is competitive, eviction with smaller budgets (2k or 4k) offers a similar trade-off in low-memory regimes. 
This suggests that large models, with their greater number of effective parameters, are more robust to the precision loss from quantization. 
However, we find that more aggressive 2-bit quantization still results in a significant loss of accuracy.

\begin{findingbox}
   KV cache eviction provides a better memory--accuracy trade-off than KV cache quantization for models with an effective size smaller than an 8-bit 8B model.
   For models at or above this threshold, quantization becomes an increasingly competitive strategy.
\end{findingbox}

\section{Related Work}
\label{sec:related_work}

\paragraph{Train-time scaling and knowledge capacity.}
Foundational scaling studies~\citep{kaplan2020scaling, henighan2020scaling, hoffmann2022training} establish power-law relationships between model size, data, and loss, yielding prescriptions for compute-optimal training under fixed compute budgets.
While these results provide guidance for allocating parameters and tokens during pre-training, they do not consider inference-time compute and hence require new extrapolations~\citep{gadre2025language}.
In parallel, capacity-oriented analyses estimate what models can store, either by modeling knowledge as information per parameter or by measuring memorization versus generalization~\citep{morris2025much,allen2024physics}. 
These views motivate a budget-centric view but leave precision and inference-time trade-offs under deployment constraints unspecified.
Bit-normalized studies examine how performance at different precisions scales with total model bits~\citep{dettmers2023case} or the amount of training data~\citep{kumar2024scaling}, particularly in zero-shot or few-shot scenarios.
\citet{feng2025numerical,mekala2025does} further show that reduced numerical precision can markedly impair arithmetic reasoning and long-context performance unless compensated by a larger model size, indicating interactions between precision, task structure, and context length.

\paragraph{Inference-time methods and scaling laws.}
Chain-of-thought prompting elicits intermediate steps, and self-consistency improves performance by sampling diverse rationales and aggregating them via majority voting~\citep{wei2022chain,wang2022self}.
Modern reasoning models are trained to generate substantially more tokens, yielding significant gains across benchmarks~\citep{wang2022self,wei2022chain,brown2024large,muennighoff2025s1,guo2025deepseek,yang2025qwen3,comanici2025gemini,jaech2024openai,qwq32b,team2025kimi}.
Test-time scaling laws study how performance changes with increased FLOPs, tokens, or the number of generations, comparing strategies such as majority voting, best-of-n, and verification-based search~\citep{brown2024large,wu2024inference,snell2024scaling,muennighoff2025s1,sadhukhan2025kinetics,wang2025scaling,zhao2025test}.
However, these studies do not capture the impact of compression techniques such as weight-only quantization, which reduces memory and latency without affecting FLOPs.
While concurrent works by~\citet{liu2025quantization} and~\citet{kurtic2025deployment} study quantization in reasoning models, their focus is on accuracy degradation rather than memory trade-offs. 
Our work is distinct in its memory-centric view: we analyze the trade-offs in allocating a fixed memory budget between model weights and test-time compute (generation length and parallelism), incorporating the full cost of the \gls{kv} cache. 
We also broaden the scope of compression techniques to include \gls{kv} cache eviction.

\paragraph{Efficient inference.}
Various strategies have been proposed to address challenges in \gls{llm} quantization, particularly for handling outliers~\citep{frantar2022gptq,xiao2023smoothquant,lin2024awq,kim2023squeezellm,dettmers2022gpt3}.
Quantization-aware training extends this idea by training models with quantized weights in the forward pass~\citep{liu2023llm,ma2024era,liu2025paretoq}.
Post-training \gls{kv} cache compression techniques can be categorized into eviction and quantization.
Eviction methods selectively discard less important entries based on different criteria~\citep{cai2025r,xiao2023efficient,zhang2023h2o,liu2023scissorhands,li2024snapkv,ge2023model}, while quantization approaches reduce the precision of cached values~\citep{badri2023hqq,kang2024gear,liu2024kivi,kim2024lexico,hooper2024kvquant}.

\section{Conclusion}
\label{sec:conclusion}

Under real-world circumstances with fixed memory budgets, deploying reasoning models is ultimately a problem of \textit{where} to spend bytes, and practitioners are presented with a myriad of choices.
Our work reformulates test-time scaling around this constraint.
We study the trade-offs in allocating memory across model size, weight precision, \gls{kv} cache compression, token budget, and sampling group size for reasoning models. 
We find that the memory-optimal inference strategy for reasoning models \textit{cannot be a one-size-fits-all prescription}: instead, it depends on the model's capacity (determined by effective size) and the nature of the task. 

For smaller model sizes (typically models under 8B), prioritizing model weights yields better memory--accuracy trade-offs by using higher-precision 8- or 16-bit weights for mathematical reasoning and favoring \gls{kv} cache eviction over quantization.
For larger models, increasing the token budget until saturation and leveraging parallel scaling become the dominant strategies. 
Importantly, the inflection point where extra \gls{kv} cache beats extra model weight may change as models become more sophisticated. 
However, by shifting the focus from FLOPs-based test-time scaling laws to practical memory constraints, our framework and analysis provide general principles for deploying reasoning models effectively.

\section{Limitations and Future Work}
\label{sec:future_work}

Our scope is intentionally focused to keep the search space tractable and inference-only.
For test-time scaling, we rely on prompt injection for serial scaling and majority voting for parallel scaling, deliberately excluding methods that require external models. 
Our analysis also does not include a comparative study of different post-training quantization or \gls{kv} cache eviction algorithms, including those requiring model retraining, such as quantization-aware training. 
We evaluate on the Qwen3 family, chosen for its broad size range and fixed architecture, and two challenging benchmarks (AIME25 for mathematical reasoning and GPQA-Diamond for knowledge-intensive reasoning).
These choices were necessary to maintain a tractable search space, which already spans more than 1,700 experimental configurations, and to focus on self-contained inference strategies, leaving a broader comparison of methods as a clear avenue for future work.

\newpage

\bibliography{iclr2026_conference}
\bibliographystyle{iclr2026_conference}

\newpage

\appendix

\section{Memory Equations}
\label{sec:memory_equations}

The total memory cost $M$ is the sum of the memory required for the model weights $M_{\mathrm{weights}}$ and the \gls{kv} cache $M_{\mathrm{kv}}$.

\paragraph{Weight Memory.}
The total memory footprint for weights is the sum of memory for the quantized and unquantized parameters. The general equation is
\[
M_{\mathrm{weights}} \approx \underbrace{\left( N_{\mathrm{quant}} \cdot \frac{P_W}{8} + \frac{N_{\mathrm{quant}}}{g_W} \cdot \frac{P_S + P_Z}{8} \right)}_{\text{Quantized Parameters}} + \underbrace{\left( N_{\mathrm{unquant}} \cdot \frac{P_{\mathrm{native}}}{8} \right)}_{\text{Unquantized Parameters}} \quad \text{[bytes]}
\]
where $N_{\mathrm{quant}}$ and $N_{\mathrm{unquant}}$ are the number of quantized and unquantized parameters, respectively, $P_W$ is the low-bit precision for weights, $g_W$ is the group size, $P_S$ and $P_Z$ are the bit-widths for the scales and zero-points, and $P_{\mathrm{native}}$ is the native precision of the unquantized layers.

In our specific setup using GPTQ, the large linear layers are quantized, while components such as the token embedding matrix, normalization layer weights, and the final language model head remain in native BF16 precision.
For our experiments, we use a group size $g_W=128$, a scale precision of $P_S=16$ (FP16), and symmetric quantization, making the zero-point precision $P_Z=0$.

\paragraph{\gls{kv} Cache Memory.}
Without compression, the \gls{kv} cache memory is given by
\[
M_{\mathrm{kv}} = G \cdot T \cdot n_{\mathrm{layers}} \cdot n_{\mathrm{kv\_heads}} \cdot d_{\mathrm{head}} \cdot 2 \cdot \frac{P_{\mathrm{native}}}{8} \quad \text{[bytes]}
\]
where $G$ is the sampling group size, $T$ is the number of tokens, $n_{\mathrm{layers}}$ is the number of layers, $n_{\mathrm{kv\_heads}}$ is the number of Key/Value heads, $d_{\mathrm{head}}$ is the dimension per head, the factor of 2 accounts for both Key and Value tensors, and $P_{\mathrm{native}}$ is the native precision of the cache elements in bits (e.g., 16 for BF16).

The memory cost is modified by different \gls{kv} cache strategies:
\begin{itemize}
    \item \textbf{Eviction:} This strategy reduces the number of tokens stored. The memory cost is
    \[
    M_{\mathrm{kv}} = G \cdot \min(T, T_{\mathrm{retain}}) \cdot n_{\mathrm{layers}} \cdot n_{\mathrm{kv\_heads}} \cdot d_{\mathrm{head}} \cdot 2 \cdot \frac{P_{\mathrm{native}}}{8}
    \]
    where $T_{\mathrm{retain}}$ is the maximum number of tokens retained by the policy. In our experiments, we use R-KV and test $T_{\mathrm{retain}} \in \{8192, 4096, 2048\}$.

    \item \textbf{Quantization:} This strategy reduces the precision but introduces overhead for quantization parameters. The cost is
    \[
    M_{\mathrm{kv}} = \left( G \cdot T \cdot n_{\mathrm{layers}} \cdot n_{\mathrm{kv\_heads}} \cdot d_{\mathrm{head}} \cdot 2 \right) \cdot \left( \frac{P_{\mathrm{kv}}}{8} + \frac{1}{g_{\mathrm{kv}}} \frac{P_S + P_Z}{8} \right)
    \]
    where $g_{\mathrm{kv}}$ is the group size, and $P_S$ and $P_Z$ are the precisions of the scales and zero-points. For our experiments, we use symmetric quantization ($P_Z=0$) with $g_{\mathrm{kv}}=64$, $P_S=16$, and test $P_{\mathrm{kv}} \in \{8, 4, 2\}$.
\end{itemize}

Below are the architectural details and per-token KV cache sizes for the evaluated models (Table~\ref{tab:model_arch_kv}).

\begin{table}[h]
\centering
\caption{\textbf{Architectural specifications and KV cache memory per token.}}
\label{tab:model_arch_kv}
\small
\begin{tabular}{l|ccc|c}
\toprule
\textbf{Model} & \textbf{$n_{\mathrm{layers}}$} & \textbf{$n_{\mathrm{kv\_heads}}$} & \textbf{$d_{\mathrm{head}}$} & \textbf{KV Cache (KB/token)} \\
\midrule
Qwen3-0.6B & 28 & 8 & 128 & 112 \\
Qwen3-1.7B & 28 & 8 & 128 & 112 \\
Qwen3-4B & 36 & 8 & 128 & 144 \\
Qwen3-8B & 36 & 8 & 128 & 144 \\
Qwen3-14B & 40 & 8 & 128 & 160 \\
Qwen3-32B & 64 & 8 & 128 & 256 \\
\bottomrule
\end{tabular}
\end{table}

\newpage

\section{Detailed Results for Parallel Scaling}
\label{sec:detailed_results_for_parallel_scaling}

\begin{figure}[h]
\begin{center}
\includegraphics[width=\linewidth]{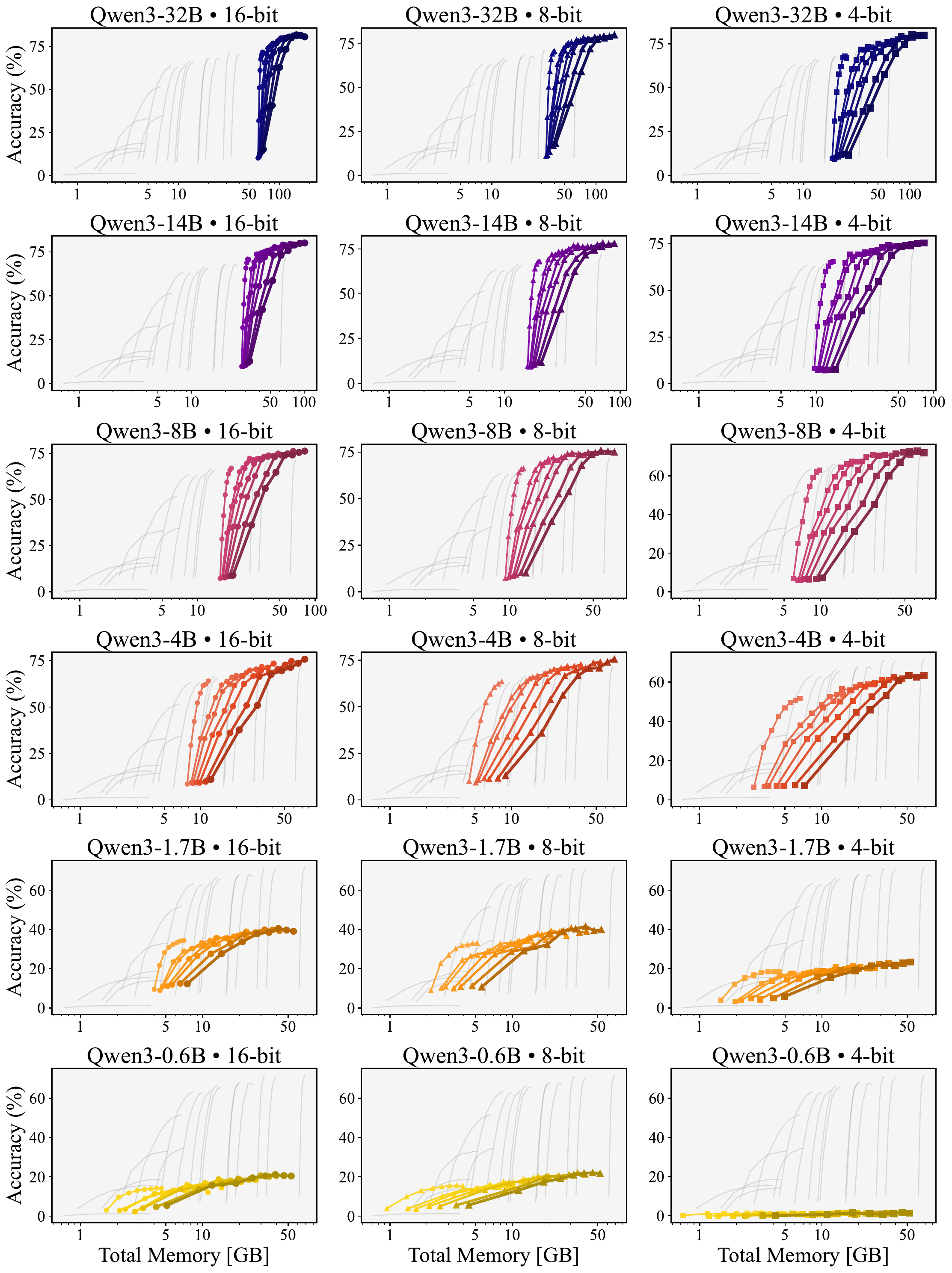}
\end{center}
\caption{
\textbf{Per-model Memory vs. Accuracy for parallel scaling (AIME25).}
Each plot shows the memory--accuracy trade-off for a single model and weight precision, comparing serial scaling ($G=1$) with parallel scaling by increasing the sampling group size, $G \in \{1, 3, 4, 6, 8, 12, 16\}$. 
Points along each curve represent increasing the token budget via budget forcing.
Parallel scaling improves the memory--accuracy trade-off only for models effectively larger than 8-bit 4B.
}
\label{fig:aime25_qwen3_parallel_scaling_individual}
\end{figure}

Figure~\ref{fig:aime25_qwen3_parallel_scaling_individual} presents the per-model plots for the parallel scaling analysis discussed in Section~\ref{sec:test_time_scaling_with_weight_only_quantization}.

\section{Detailed Results for KV Cache Compression}
\label{sec:detailed_results_for_kv_cache_compression}

\begin{figure}[h]
\begin{center}
\includegraphics[width=\linewidth]{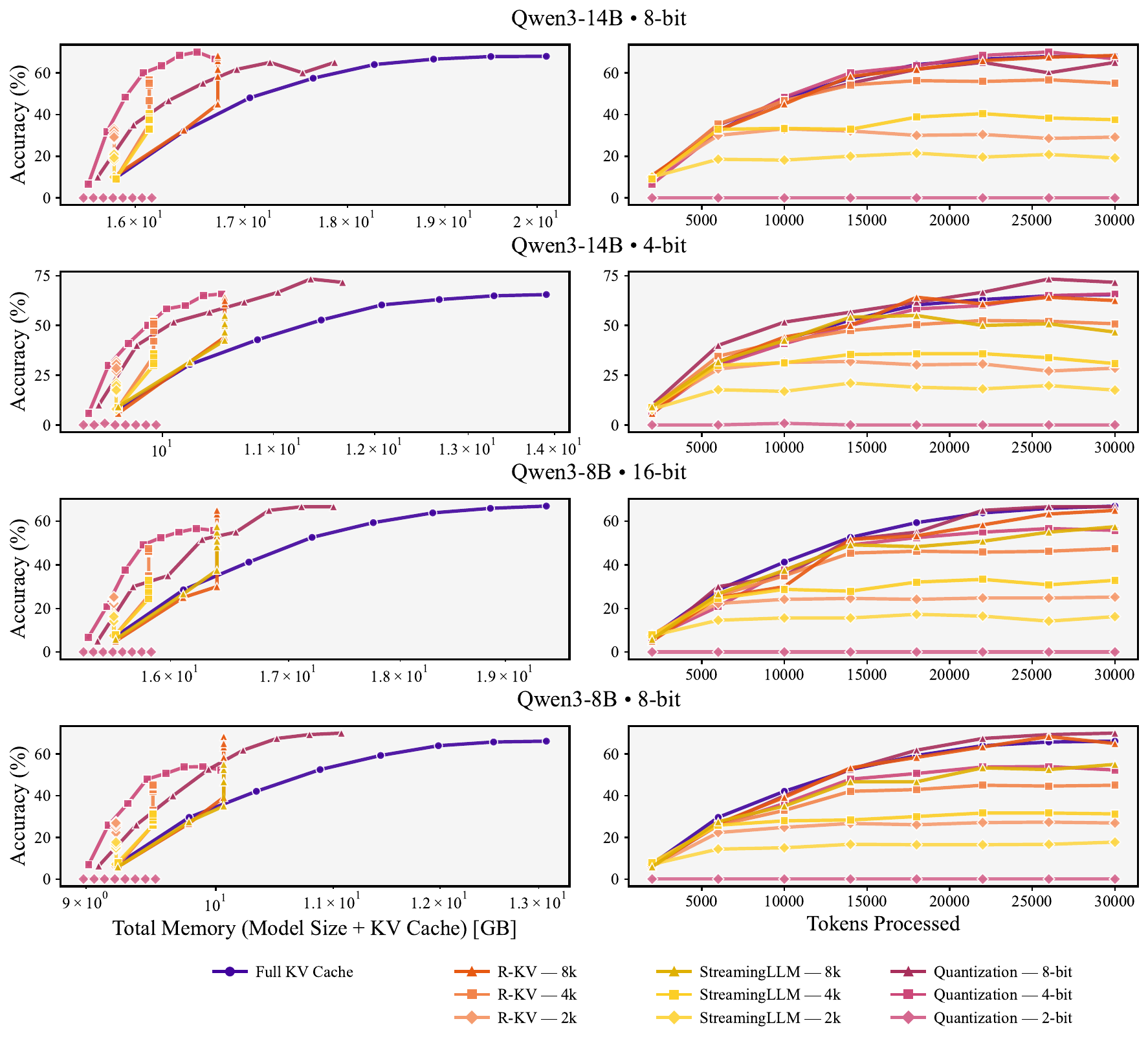}
\end{center}
\caption{
\textbf{Per-model Memory vs. Accuracy by \gls{kv} cache strategy (AIME25, models $>$ 4B).}
Each plot shows the memory--accuracy trade-off for a single model and weight precision, comparing KV cache eviction methods (R-KV, StreamingLLM) against KV cache quantization and no compression.
Points along each curve represent increasing the number of processed tokens.
}
\label{fig:aime25_qwen3_kv_strategy_all_configs_part1}
\end{figure}

\begin{figure}[h]
\begin{center}
  \includegraphics[width=\linewidth]{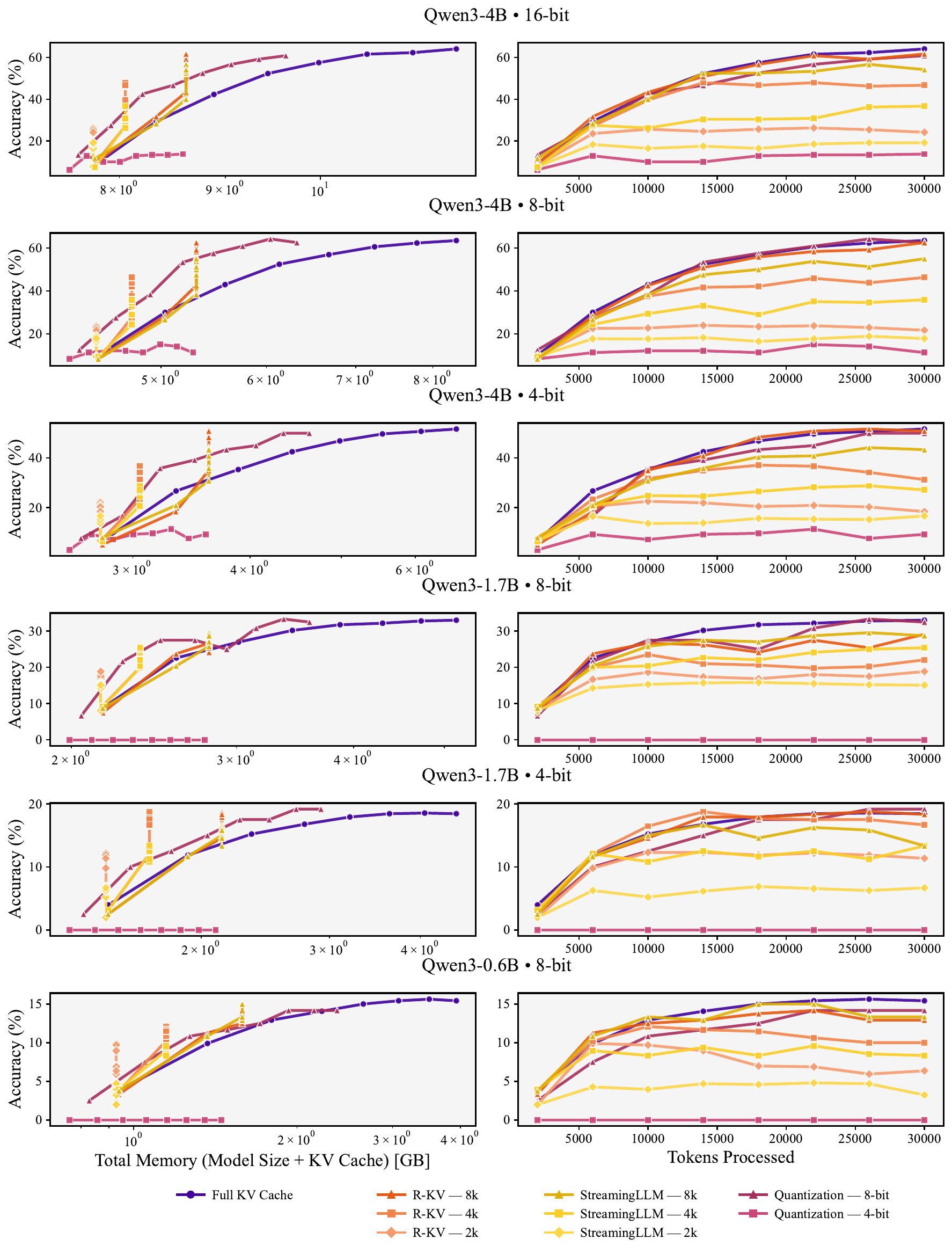}
\end{center}
\caption{
\textbf{Per-model Memory vs. Accuracy by \gls{kv} cache strategy (AIME25, models $\le$ 4B).}
Each plot shows the memory--accuracy trade-off for a single model and weight precision, comparing KV cache eviction methods (R-KV, StreamingLLM) against KV cache quantization and no compression.
Points along each curve represent increasing the number of processed tokens.
}
\label{fig:aime25_qwen3_kv_strategy_all_configs_part2}
\end{figure}

Figures~\ref{fig:aime25_qwen3_kv_strategy_all_configs_part1} and \ref{fig:aime25_qwen3_kv_strategy_all_configs_part2} show the per-model results for the \gls{kv} cache compression analysis discussed in Section~\ref{sec:test_time_scaling_with_weight_and_kv_cache_compression}.
For eviction, we also present results for StreamingLLM, where we retain the first $T_{\text{retain}}/2$ tokens and the most recent $T_{\text{retain}}/2$ tokens for a given retention budget $T_{\text{retain}}$.

\end{document}